\documentclass[letterpaper, 10 pt, conference]{ieeeconf}  

\IEEEoverridecommandlockouts                              

\usepackage{amsmath}
\usepackage{amssymb}

\usepackage{tabularx}
\usepackage{graphicx}
\usepackage[font=footnotesize]{caption}
\usepackage{color}

\usepackage{subfig}

\newcommand{\vecb}[1]{\ensuremath{\mathbf{#1}}}

\title{\LARGE \bf
Contact Optimization for Non-Prehensile Loco-Manipulation via Hierarchical Model Predictive Control
}

\author{Alberto Rigo, Yiyu Chen, Satyandra K. Gupta, and Quan Nguyen
\thanks{$^{1}$Alberto Rigo, Yiyu Chen, Satyandra K. Gupta, and Quan Nguyen, are with the Department of Aerospace and Mechanical Engineering,
        University of Southern California, Los Angeles, CA, 90089
        {\tt\small rigo@usc.edu}, {\tt\small yiyuc@usc.edu}, {\tt\small quann@usc.edu}, {\tt\small guptask@usc.edu}}%
}

\begin{document}

\maketitle
\thispagestyle{empty}
\pagestyle{empty}

\begin{abstract}
Recent studies on quadruped robots have focused on either locomotion or mobile manipulation using a robotic arm.
Legged robots can manipulate heavier and larger objects using non-prehensile manipulation primitives, such as planar pushing, to drive the object to the desired location. In this paper, we present a novel hierarchical model predictive control (MPC) for contact optimization of the manipulation task. 
Using two cascading MPCs, we split the loco-manipulation problem into two parts: the first to optimize both contact force and contact location between the robot and the object, and the second to regulate the desired interaction force through the robot locomotion. 
Our method is successfully validated in both simulation and hardware experiments. While the baseline locomotion MPC fails to follow the desired trajectory of the object, our proposed approach can effectively control both object's position and orientation with a minimal tracking error. 
This capability also allows us to perform obstacle avoidance for both the robot and the object during the loco-manipulation task.
\end{abstract}

\section{INTRODUCTION}
Legged robots have great potential to interact with the environment and have demonstrated significant performance for locomotion, such as high-speed running and robust walking on challenging terrains \cite{di2018dynamic, sombolestan2021adaptive, jenelten2020perceptive, miki2022learning, kim2019highly, nguyen20151, nguyen2020dynamic, nguyen2018dynamic}. With the existing control and planning algorithms, most applications for quadruped robots focus on navigation and inspection which always try to avoid objects/obstacles even if they are movable \cite{grey2017footstep, cebe2021online, dai2014whole}. 
In this paper instead, we are interested in realizing the capability of legged robots leveraging their body during locomotion to manipulate a heavy object. 

In mobile manipulation, robots can exhibit different modes of interaction with the object. For example, mobile robots equipped with a robotic arm \cite{xin2022loco, ferrolho2022roloma, rehman2016towards, bellicoso2019alma, sleiman2021unified} can enable basic manipulation tasks such as door opening, pick-and-place, and load carrying. 
However, such setups are limited to small payload and object dimension due to the payload limit of the portable robot arm. 
For legged robots, manipulation with their feet is also an intriguing idea as quadrupedal animals can use their legs or limbs for manipulation\cite{shi2021circus, ji2022hierarchical}. However, this setup is unsuitable for loco-manipulation tasks, which require the robot to move and manipulate the object simultaneously because it requires both locomotion and manipulation. 
If one of the legs is used for manipulation, the locomotion task will become challenging for quadruped robots. 
Therefore, this paper tackles the problem of loco-manipulation for quadruped robots using a planar pushing motion.
For large and heavy objects, non-prehensile manipulation such as planar pushing offers excellent advantages. When the object is too large or too heavy to be grasped, pushing becomes one of the options to drive it to the desired state. 
In addition, this method also allows quadruped robots to manipulate objects without adding an additional robotic arm.

\begin{figure}
    \centering
    \includegraphics[width = \linewidth]{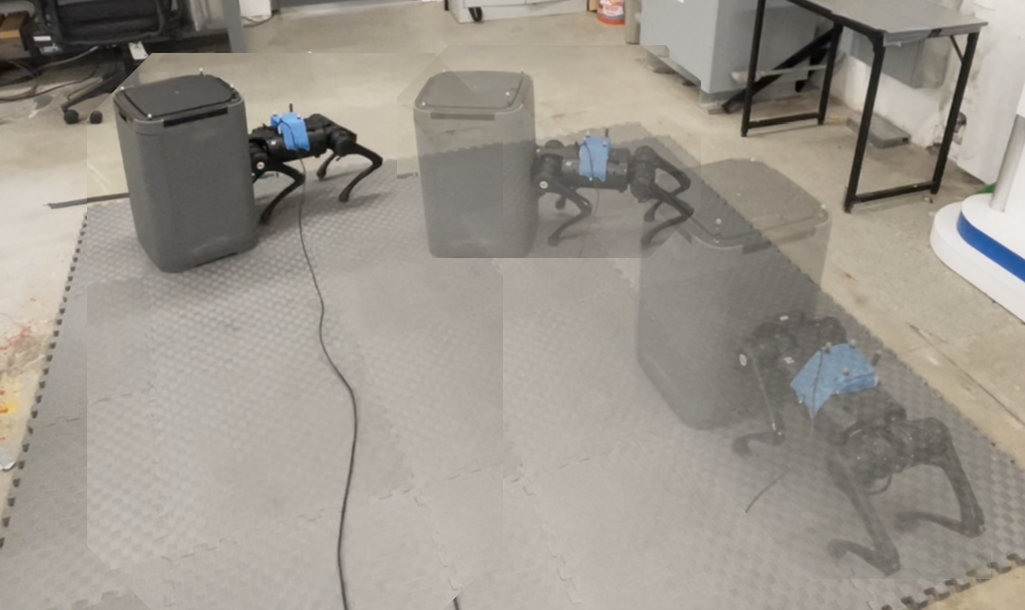}
    \caption{Motion snapshots of Unitree A1 robot manipulating a $5~kg$ object to follow a circular trajectory.}
    \label{fig:obj_robot}
\end{figure}

Pushing is a widely used motion primitive and has been thoroughly studied by the manipulation community. The mechanics of planar pushing is well-studied in \cite{mason1986mechanics, akella1998posing, zhou2016convex}. Some motion planning algorithms \cite{lynch1996stable, zhou2017fast} are introduced to find open-loop trajectories to drive the object to the target pose, assuming that the manipulator always sticks with the object for the entirety of the push. To handle the complexity associated with frictional contact interactions, motion planning algorithms developed by the robotic manipulation community manage to handle different mode sequences \cite{hou2018fast, woodruff2017planning, toussaint2018differentiable}. Nevertheless, these approaches are computationally heavy due to the nonlinear and non-convex optimization programs. 
A recent work in \cite{hogan2020reactive} proposes a real-time controller to reason across different contact modes, including sticking and sliding, using an online approximation for the offline mix-integer program. 

The recent developments on model predictive control for legged robot locomotion \cite{li2021force, di2018dynamic, sleiman2021unified} suggest that optimal control action can be computed online given a proper contact schedule. However, these works mainly focus on locomotion. To simultaneously achieve locomotion and manipulation tasks, we propose a novel hierarchical MPC framework including (1) high-level manipulation MPC to optimize for both contact force and contact location of the manipulation task; and (2) low-level loco-manipulation MPC to regulate the interaction force between the robot and the object while maintaining the desired locomotion performance. 
Both MPC problems are solved effectively in real-time.
Numerical and experimental validation have shown that our approach outperform locomotion MPC or heuristic approach for loco-manipulation.  
Thanks to the capability of optimizing contact location, our approach can allow legged robots to manipulate heavy objects effectively with a highly accurate position and orientation tracking. This also enables the execution of collision-free trajectory for both the robot and the object. 



The rest of the paper is organized as follows. Section \ref{sec:2} introduces the object-robot system for non-prehensile body loco-manipulation. Section \ref{sec:3} presents the proposed control architecture and the two MPC in detail. Then, Section \ref{sec:4} shows simulation and hardware experiments results. Finally, Section \ref{sec:5} draws conclusion remarks.
\section{SYSTEM OVERVIEW}
\label{sec:2}

In this paper, we are interested in pushing an arbitrary object, following a planned trajectory in terms of $x$ and $y$ world frame position and heading angle $\psi$. We assume we know all the geometric and inertial characteristics of the object, and we have the feedback on its heading angle and center of mass position. Due to the limitations of the pushing primitive, to move the object to the desired location, we have to align its heading angle toward that location. Leveraging the position tracking of the quadruped robot, we can optimize the contact point between the robot head and object to push forward and, at the same time, rotate the object to align the heading angle to the desired one. Without changing the contact location, we would not be able to control the heading angle of the object.
The nonlinearity of the loco-manipulation problem is solved by splitting it into two separate linear parts, the first responsible for determining the required manipulation action to be exerted on the object; the second responsible for the locomotion under the effect of the contact interaction.

\section{PROPOSED FRAMEWORK}
\label{sec:3}

\begin{figure}[tp]
    \medskip
    \centering
    \includegraphics[width = \linewidth]{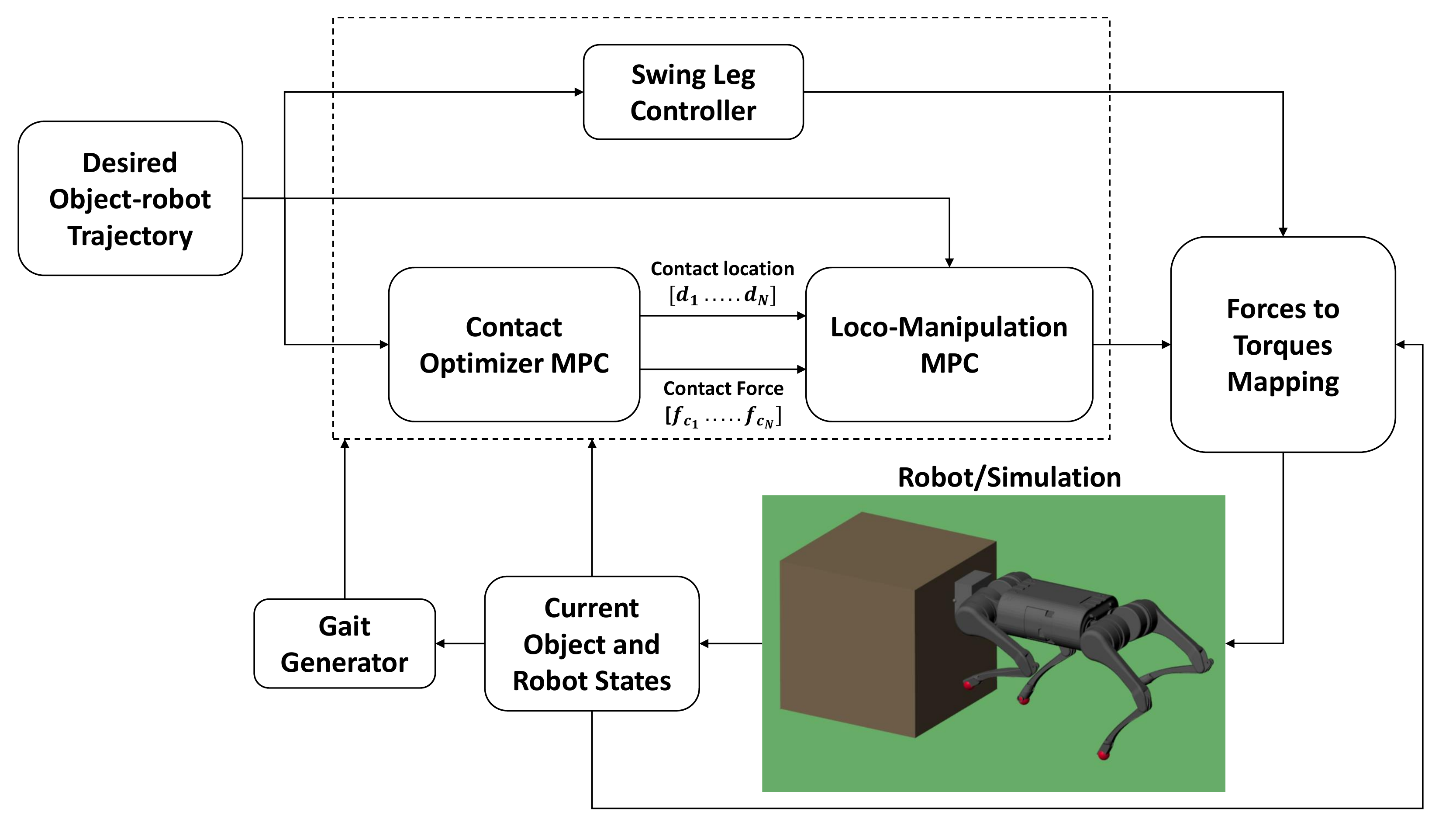}
    \caption{Control Architecture}
    \label{fig:control_arch}
\end{figure}

The high-level control comprises the swing leg controller and two Model Predictive Controllers (MPC) in a hierarchical structure, as depicted in Fig. \ref{fig:control_arch}. First, the contact optimizer MPC is used to compute the required control input, i.e., contact force and contact point on the object surface, to drive the manipulated object to the desired states. Then, the loco-manipulation MPC is responsible for tracking the planned trajectory for the object-robot system based on the output of the contact optimizer MPC. The two MPCs use the same prediction horizon, so the predicted values for the contact interaction by the contact optimizer MPC are used as inputs for the loco-manipulation MPC.

\subsection{Contact Optimizer MPC}
\label{subsec:contact_MPC}

\begin{figure}[tp]
    \medskip
    \centering
    \includegraphics[width = \linewidth]{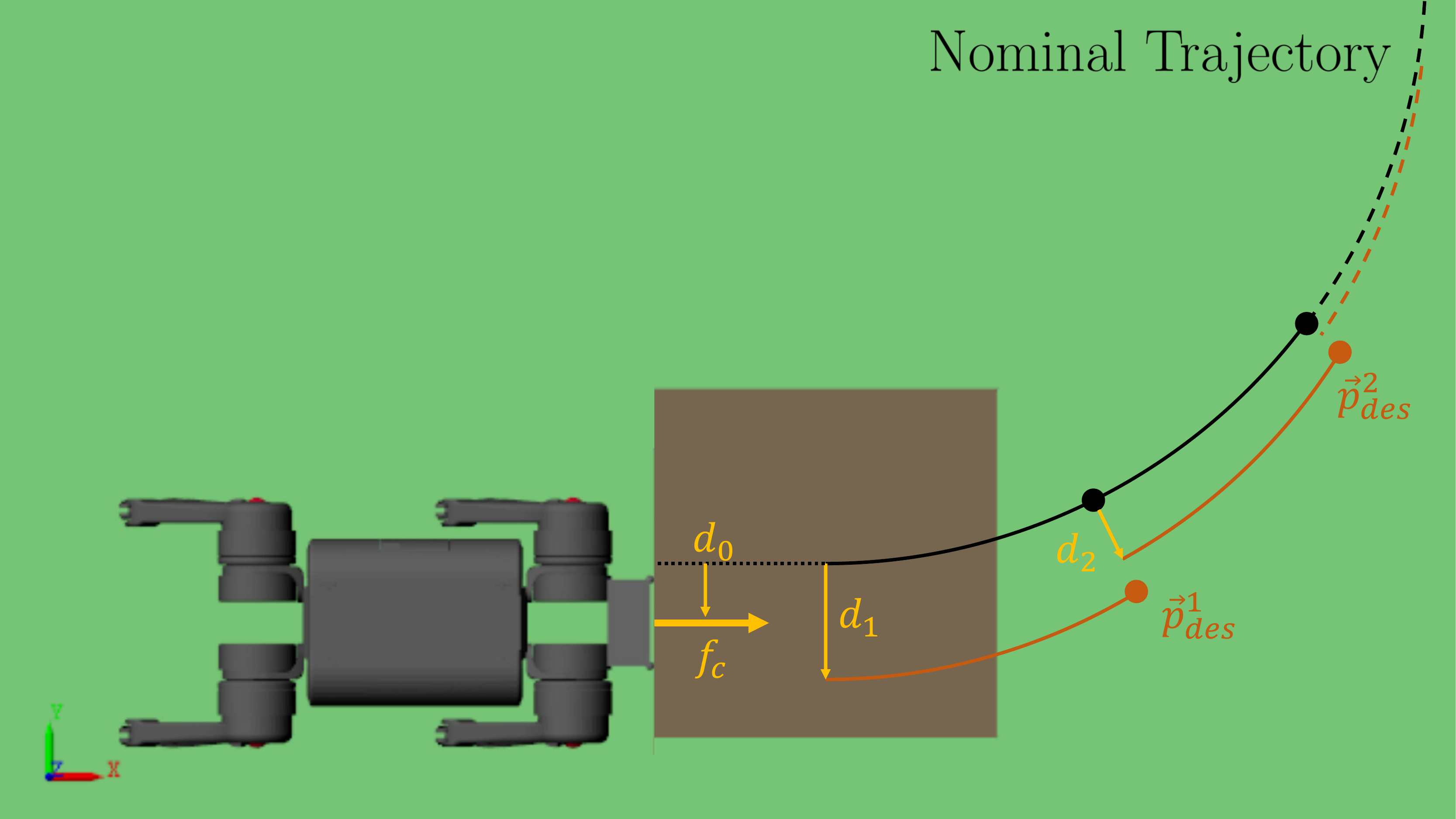}
    \caption{Object-robot system overview and representation of the offset trajectory derived from the contact point optimization}
    \label{fig:ref_traj}
\end{figure}

This first controller uses a simplified model of manipulated object dynamics. In this paper, we are interested in controlling the object's position and heading angle. Therefore we can use the following simplified rigid body dynamics equations:
\begin{align}
    m\Ddot{\vecb{p}}_{obj} &= \vecb{f}_{\mu} + \vecb{f}_c\\
    I_z \Dot{\omega}_z &= \vecb{f}_c\times \vecb{d}
    \label{eq:obj_dyn}
\end{align}
where $\vecb{p}$ represents the position of the object in the world frame, $\vecb{f}_{\mu}$ is the frictional force between object and ground, and $\vecb{f}_c$ is the contact force applied to the object by the robot in world frame, $\omega_z$ is the angular velocity of the object in the vertical direction with respect to its center of gravity, and $\vecb{d}$ is the vector between the contact point and object center of mass in world frame. If we consider the contact force and the contact point as control variables for the problem, the previous set of equations is nonlinear. We can make some assumptions and simplifications to use them as model dynamics in a linear MPC. First, with a small enough MPC frequency update, we can assume that the contact force will change only by a small amount; hence the contact force used in the eq. (\ref{eq:obj_dyn}) is the known value of force computed at the previous controller update, $f_{c_0}$. Then, we can further assume the contact force $\vecb{f}_c$ is always in the x-direction of the object body frame, simplifying the definition of the contact point $\vecb{d}$. It becomes the distance from the center of mass of the object in the y direction of the object body frame, as seen in Fig \ref{fig:ref_traj}. With these assumptions, equations (\ref{eq:obj_dyn}) are now linear and can be used to represent the object dynamics in the state space form:
\begin{equation}
    \Dot{x} = Ax + Bu
    \label{eq:dyn_constr}
\end{equation}
where $x = \begin{bmatrix}\psi & x & y& \omega_z & \Dot{x} & \Dot{y} & g\end{bmatrix}$, $u = \begin{bmatrix}f_c & d\end{bmatrix}$, and the matrices are
\begin{equation}
    A = \begin{bmatrix}\vecb{0}_{3\times3} & \vecb{I}_{3\times3} & \vecb{0}_{3\times1} \\ \vecb{0}_{3\times3} & \vecb{0}_{3\times3} & \begin{bmatrix}0\\-\mu\\-\mu\end{bmatrix}\\\vecb{0}_{1\times3} & \vecb{0}_{1\times3} & 0 \end{bmatrix}, \quad  B = \begin{bmatrix} \vecb{0}_{3\times2}\\\begin{bmatrix}f_{c_0} & 0\\0 & \cos{\psi} \\ 0 & \sin{\psi}\end{bmatrix}\\\vecb{0}_{1\times2}\end{bmatrix}.
    \label{eq:dyn_matrices}
\end{equation}
where we assumed that the frictional force $\vecb{f}_{\mu}$ is expressed as $-\mu m g$ for both $x$ and $y$ directions, the body frame contact force $f_c$ can be expressed in world frame using a rotation matrix $\vecb{R}_{\psi} \in \mathbb{R}^2$, and $f_{c_0}$ is the contact force computed at the previous controller update.

We discretize the dynamics model equations, and using a classic quadratic programming (QP) formulation, we can solve the MPC problem with $N$ horizons and a cost function
\begin{equation}
    \min_{\vecb{x},\vecb{u}}  \sum_{i=1}^{N}(\vecb{x}_{i+1}-\vecb{x}_{{i+1}_{ref}})^T \vecb{Q} (\vecb{x}_{i+1}-\vecb{x}_{{i+1}_{ref}}) + ||\vecb{u}_i|| \vecb{R},
    \label{eq:cost}
\end{equation}
where we want to minimize the difference between the object's current and reference states and the control effort. The reference states of the object are computed based on the desired trajectory we want it to follow. The MPC controller solves for the optimal contact force $f_c$ and contact point $d$ with respect to dynamic constraints (\ref{eq:dyn_constr}) and the following inequality constraints:
\begin{align}
    d_{min} < d < d_{max} \\
    0 < f_c < F_{max}
\end{align}
Here, the first constraint represents limits based on the dimension of the object, while the second one is used to maintain contact and pushing action between object and robot and avoid requiring an instantaneous force too large to ensure the stability of the locomotion.
The optimized result of this MPC alters the loco-manipulation MPC in two ways, explained in the next section.

\subsection{Loco-manipulation MPC}
To effectively regulate the desired contact force $f_c$ and contact location $d$ derived from the contact optimizer MPC in Section \ref{subsec:contact_MPC}, we present a unified loco-manipulation MPC that takes into account these two variables in the control design.
In comparison with the locomotion MPC \cite{di2018dynamic}, following are the main developments of our framework for loco-manipulation.

\begin{itemize}
    \item 
    Our loco-manipulation MPC takes into account the interaction force $f_c$ between the robot and the object in the robot dynamics. Therefore, it can regulate the desired manipulation force while maintaining desired performance for locomotion.
    \item 
    The reference trajectory of the robot locomotion is also automatically updated based on the desired contact location $d$ as well as real-time feedback of the object state.
\end{itemize}

The single rigid body dynamics (SRBD) equations used for MPC are modified as follows:
\begin{align}
    m\vecb{\Ddot{p}} &= \sum_{i=1}^{4}\vecb{f}_i - \vecb{f}_g - \vecb{f}_c , \\
    \frac{d}{dt}\vecb{I}\vecb{\omega} &= \sum_{i=1}^{4}(\vecb{r}_i - \vecb{p})\times\vecb{f}_i ,
\end{align}
where $m$ is the robot mass; $\vecb{p}$ and $\vecb{r}_i$ are the body position and foot position in the world frame; $\vecb{I}$ and $\vecb{\omega}$ are the rotational inertia tensor of the body and angular velocity of the body expressed in the body frame. Finally, $\vecb{f}_i$, $\vecb{f}_g$, and $\vecb{f}_c$ are the vectors for reaction forces, gravitational forces, and contact force with the object, expressed in the world frame. Here, the direction of the contact force $\vecb{f_c}$ is assumed to be constant in body frame, acting in the longitudinal direction, and needs to be properly expressed in world frame using the rotation matrix of the body. We don't include the moment generated by the contact force $\vecb{f_c}$ in the rotational dynamics equations since it's magnitude is small relative to the moment generated by the reaction forces. It would also be difficult to effectively represent, due to the uncertainty on the contact point location between robot head and object.

These equations are discretized and used in an MPC formulation with $N$ horizons and a prediction horizon representing a full gait cycle. The MPC is formulated as a quadratic program (QP) that can efficiently be solved in real-time. The cost function for the MPC problem is similar to (\ref{eq:cost}). However, in this case, $x$ represents the states of the robot body, and $u$ represents the reaction forces on the four feet. 

The second difference with respect to a conventional locomotion MPC for legged robots lies in the definition of the reference states. We take the nominal reference trajectory for the object-robot system, and we offset it by the contact point $d$ computed in the contact optimizer MPC. The changes occur only in the $x-y$ plane trajectory definition, as shown in fig \ref{fig:ref_traj}. We use the same prediction horizon for the two MPCs to offset the trajectory by the optimal distance $d_i$ for each horizon and we set the desired yaw angle of the robot to be equal to the box heading angle. Since we are tracking a relative position between the robot and the box, the position gains in the matrix $\vec{Q}$ of the cost function \ref{eq:cost} are set to a relatively high value, to ensure the capability in tracking the contact point location.

\section{RESULTS}
\label{sec:4}
This section presents the validation and results of simulation and hardware experiments using a Unitree A1 robot. For simplification purposes, we show results using a box of mass $m$ and known dimensions, but as we showed in the previous section, the proposed framework is generalizable to an arbitrary object.

\subsection{Simulation}
The simulation environment used is the Matlab Simscape Multibody package, able to represent the interaction between robot, object, and ground plane. Both MPCs prediction horizon is set to $30~ms$ with a sampling frequency of $3~ms$, for a total of 10 horizons, in line with other locomotion controllers. We then use the predicted optimal contact force and contact point distance for all ten horizons as inputs to the loco-manipulation MPC. The mass of the box is $5~kg$, and the coefficient of friction between the box and the ground is $0.5$.

First, we present the comparison in performing two tasks between 3 types of controllers:
\begin{itemize}
\item (a) Baseline locomotion MPC with a fixed contact location.
\item (b) Locomotion MPC + a heuristic policy to adjust the contact location 
\item (c) Our proposed controller using hierarchical MPC to optimize for both contact force and contact location.
\end{itemize}

To better emphasize the advantage of our proposed approach on contact optimization, we also investigate a heuristic policy to adjust the contact location to allow the robot to control the yaw motion of the object. The heuristic policy commands a positive or negative lateral velocity, in the robot frame, based on the heading angle direction, to properly change the contact point and adjust the box direction of motion, as follows:

\begin{equation}
    v_y = v_y^{des}\text{sign}(\psi_{box} - \psi_{target})
\end{equation}
where $\psi_{box}$ is the heading angle of the box, and $\psi_{target}$ is the heading angle from the current box position to the target. With this policy, the robot tries to align the box orientation to face the target. To have less chattering of the robot lateral motion, we used the deadband function instead of the sign function, leading to a slight decrease in tracking performance. Moreover, with this policy the robot could fail the task by moving out of the object dimensions, since there is no constraint to keep the contact within the limits, unlike in our proposed approach.
\begin{figure}[tp]
    \medskip
    \centering
    \includegraphics[width = \linewidth]{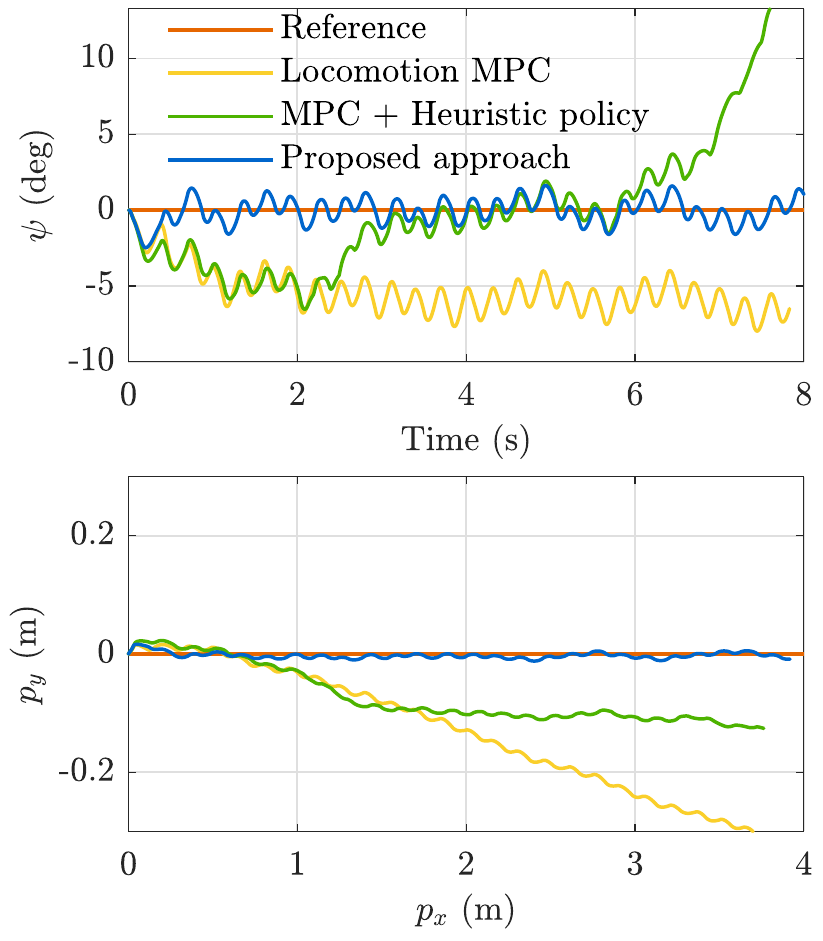}
    \caption{Comparison of box's $x-y$ plane position and heading angle following a straight line trajectory for the 3 controllers}
    \label{fig:comp_heur_straight}
\end{figure}
 
\begin{figure}
    \centering
    \includegraphics[width = \linewidth]{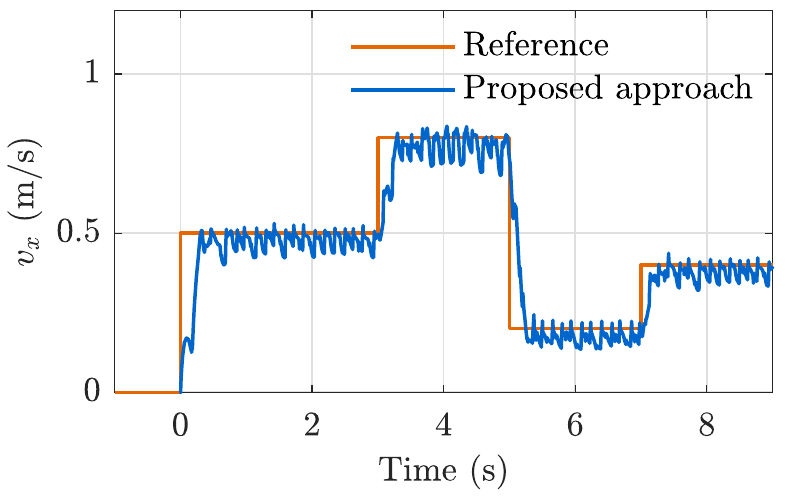}
    \caption{Tracking of the box velocity with step changes every 2 seconds to the desired velocity}
    \label{fig:comp_heur_straight_velocity}
\end{figure}

The first task is reaching the desired target location following a straight line. The results are presented in fig. \ref{fig:comp_heur_straight}. Here, we can see that the baseline locomotion MPC fails, while the other two approaches can reach the target. On the contrary, only the proposed approach can keep the robot in a straight line toward the target, thanks to the optimized contact point location updating in real-time based on the dynamics of the box. Moreover, in Fig. \ref{fig:comp_heur_straight_velocity}, we can see that the proposed controller can effectively track sudden changes in the desired velocity for the object-robot system, thanks to the optimal values of contact force computed in the contact optimizer MPC.

\begin{figure}[tp]
    \medskip
    \centering
    \includegraphics[width = \linewidth]{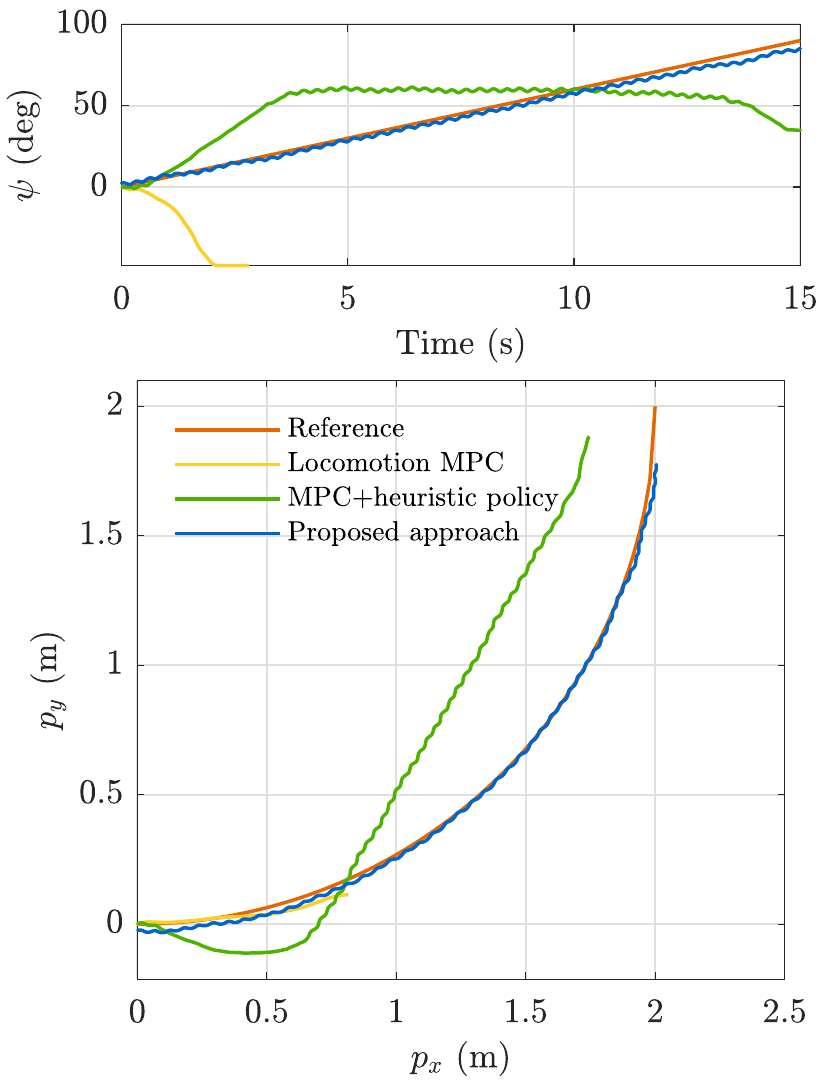}
    \caption{Comparison of box's $x-y$ plane position and heading angle following a quarter circle trajectory for the three controllers}
    \label{fig:comp_heur_curved}
\end{figure}

The next task consists in reaching the desired target while following a curved trajectory, in this case, one-quarter of a circle. The baseline locomotion MPC immediately fails, losing contact with the box since there is no policy to adjust the contact point location and this would not allow to control the heading angle. With this task, we can highlight the difference between a controller that considers the manipulated object dynamics and adjusts its action in real-time and a simple policy for changing the pushing contact point. Fig. \ref{fig:comp_heur_curved} shows the results for this task, and we can see that, while both controllers can accomplish the task, only our proposed approach can successfully follow the desired trajectory. This is because the heuristic controller has the limitations of tracking only the box's heading angle while the robot pushes it forward to reach the target. Instead, our proposed approach shows that it can dynamically change contact points to adjust the box position and orientation, always maintaining the contact location within the limits.

\begin{figure}[tp]
    \bigskip
    \centering
    \includegraphics[width = \linewidth]{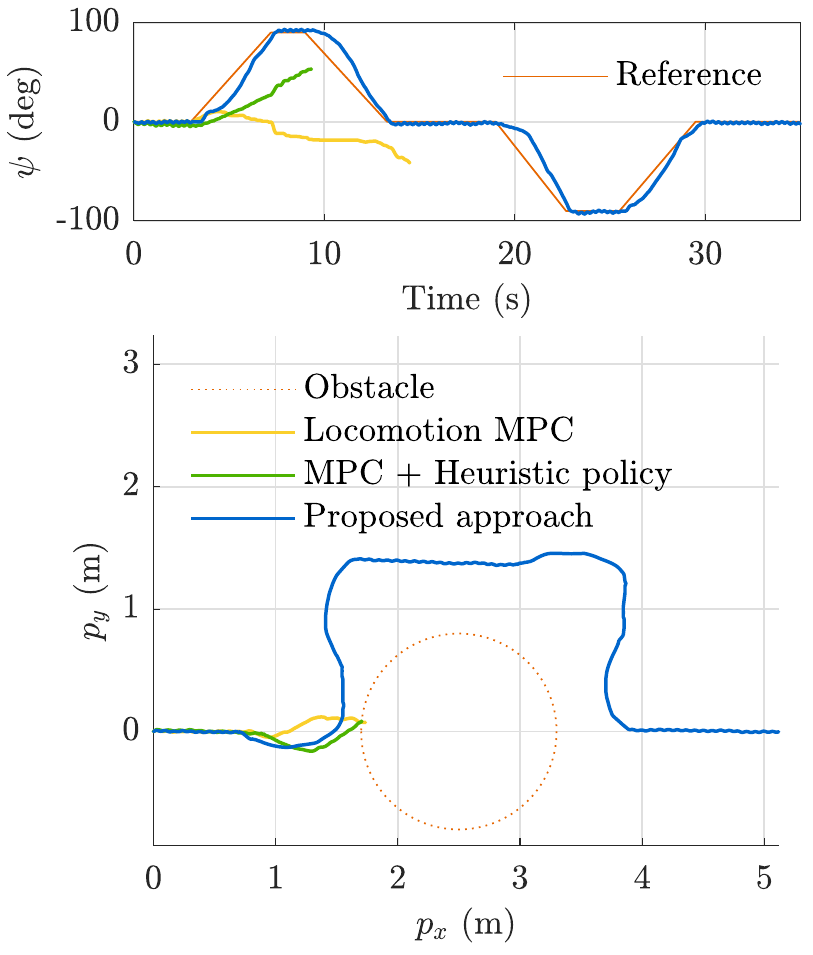}
    \caption{Following a collision-free trajectory in an obstacle avoidance scenario. Box's $x-y$ plane position and heading angle are shown.}
    \label{fig:obst_avoid}
\end{figure}

One real-world scenario where our proposed controller can be applied is pushing an object through a path while avoiding obstacles in the environment. The obstacle avoidance policy is not the focus of this paper, and we can assume the desired trajectory is already collision-free. With the following sets of plots, we want to show the capability of following sharp changes in the direction of motion, typically occurring in obstacle avoidance. Fig \ref{fig:obst_avoid} shows our controller's ability to change the contact point to react quickly to sharp changes in desired heading angle, unlike the other two controllers that fails by colliding with the obstacles. One limitation of our approach is that the desired forward velocity must be tuned down during the turns to facilitate the motion. While during straight pushing, we can track velocity larger than 0.5 m/s, during a sharp turn, we have to limit it to 0.1 m/s. 

\subsection{Hardware Experiments}
For hardware experiments, we use an Unitree A1 robot, with a low friction head, to avoid sticking between the head and the box surface. The mass of the box we use in experiments is $5~kg$, and we use an estimated coefficient of friction of 0.2. The heading angle and position feedback for both robot and box are obtained using an Optitrack motion capture system (MoCap). Since the tracking precision needed to follow the optimal contact point on the box is high, we could not rely consistently on the robot's internal state estimation, which showed drift in position estimation. The MoCap system comprises 6 Optitrack $\text{Prime}^{X}$ 13W, for improved tracking of ground objects, with a tracking frequency of $100\text{Hz}$. The trackable surface area by the MoCap for the hardware experiments is $2\times2$ meters, limiting the commanded forward speed to the robot in order to show a long enough control experiment.

\begin{figure}[tp]
    \medskip
    \centering
    \includegraphics[width = \linewidth]{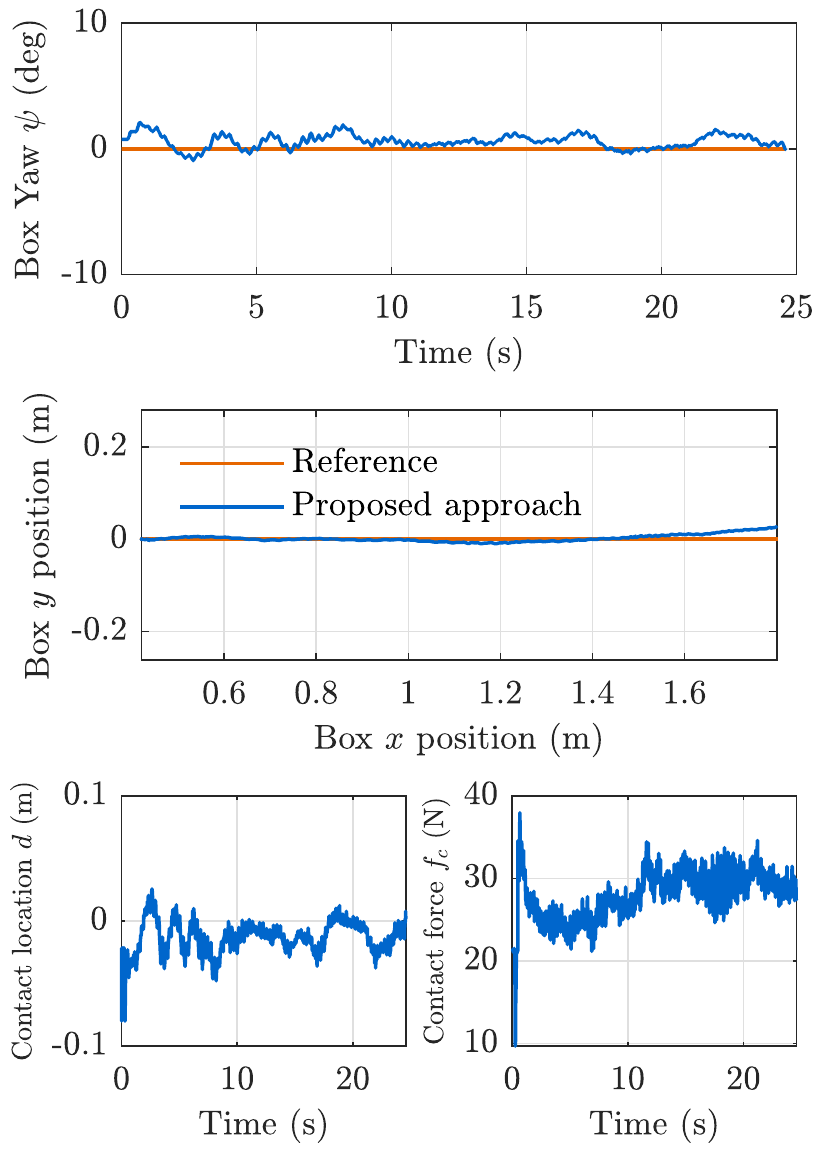}
    \caption{Experimental results obtained with MoCap system during the task of following a straight line and contact location and contact force computed by the contact optimizer MPC}
    \label{fig:exp_straight}
\end{figure}

The first task consist in following a straight line and maintaining a constant heading angle, similar to the results shown in the simulation. Fig. \ref{fig:exp_straight} shows the heading angle and $x-y$ plane position of the box. We can see that the controller can correct the direction of the box to bring it back to the desired trajectory, even with disturbances and model uncertainties present in hardware experiments. The optimized contact location is within the limits imposed by the box dimensions and oscillates around zero to adjust the heading angle during the experiment. The contact force is higher during the initial phase, when it needs to accelerate the box, and then settles to a steady state value needed to keep the box at the desired velocity.

\begin{figure}[tp]
    \medskip
    \centering
    \includegraphics[width = \linewidth]{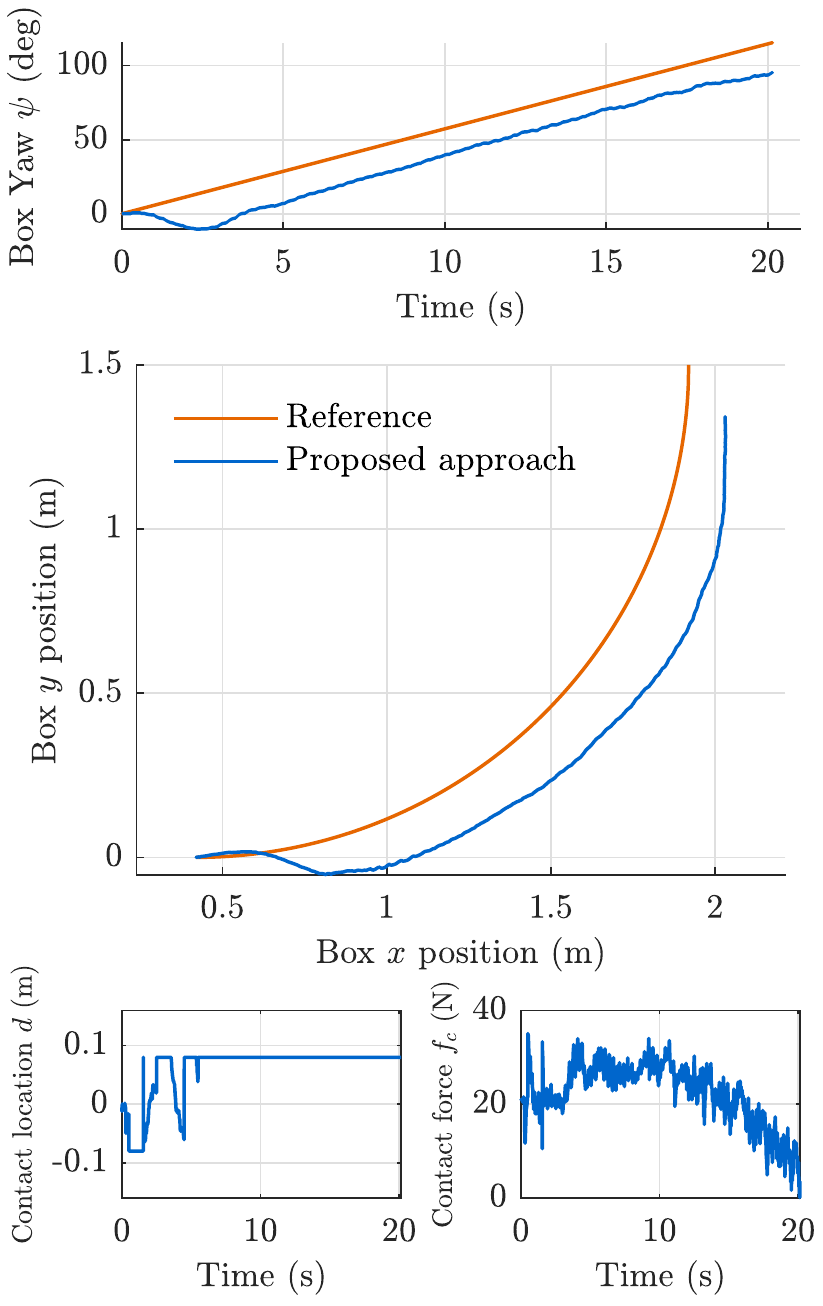}
    \caption{Experimental results obtained with MoCap system during the task of following a curved trajectory and contact location and contact force computed by the contact optimizer MPC}
    \label{fig:exp_curved}
\end{figure}

The next task is following a curved trajectory with a constant turning velocity. The results are shown in Fig. \ref{fig:exp_curved}. Due to an initial swerving off the trajectory, we have a steady state error in both position and orientation tracking due to uncertainties in the inertial and frictional quantities of the box. Still, the tracking results are meaningful, and the optimal contact point computation allows following the desired trajectory also in hardware experiments. We can see that it saturates to the maximum value imposed by the box dimensions when the controller is trying to track the offset error in the  heading angle of the box. Thanks to the predicted values of contact point in the contact optimizer MPC we can ensure that we keep the contact location within the box limits.

\section{CONCLUSIONS}
\label{sec:5}
In summary, we have presented a practical approach with a hierarchical structure comprising two MPCs, to solve the challenging problem of body loco-manipulation. The hierarchical structure allows us to simplify the nonlinear nature of the problem. We have demonstrated our approach's effectiveness using numerical and experimental validations.  In simulation, we have shown that the proposed approach improves to the loco-manipulation problem with respect to other controllers, thanks to the contact point location and contact force optimization. This allows the robot to follow diverse trajectories with changes in velocity and sharp turns, which is helpful in many scenarios. We also impose constraints on the problem to ensure that the robot is always in contact with the box, taking into account the dimensions of the object. In experiments, we have replicated the results obtained in simulation, showing the successful implementation of contact optimization. We showed our approach can effectively push a box of $5~kg$, accounting for \%50 of the robot mass. In the future, we will extend our framework to consider online obstacle avoidance through control barrier functions so that the robot can navigate an environment with obstacles while pushing an object. With this implementation we could ensure collision free trajectories with online adaptations.





\newpage
\bibliographystyle{IEEEtran}
\bibliography{IEEEexample}

\end{document}